\documentclass{article}

\usepackage{microtype}
\usepackage{graphicx}
\usepackage{subfigure}
\usepackage{booktabs} 
\usepackage{multirow}
\usepackage{hyperref}
\usepackage{arydshln}
\usepackage{colortbl}
\usepackage{subcaption}

\usepackage[accepted]{icml2024}
\usepackage{enumitem}
\usepackage{amsmath}
\usepackage{amssymb}
\usepackage{mathtools}
\usepackage{amsthm}

\usepackage[capitalize,noabbrev]{cleveref}
\usepackage{graphicx,subfigure}

\theoremstyle{plain}

\theoremstyle{definition}

\theoremstyle{remark}

\usepackage[textsize=tiny]{todonotes}

\usepackage{amsmath,amsfonts,bm}

\newcommand{\mbr}[1]{\mathbb{R}^{#1}}

\def\eqref#1{(\ref{#1})}

\def\1{\bm{1}}

\def\vx{{\bm{x}}}

\def\mF{{\bm{F}}}

\def\mM{{\bm{M}}}

\def\mU{{\bm{U}}}
\def\mV{{\bm{V}}}

\def\mX{{\bm{X}}}

\DeclareMathAlphabet{\mathsfit}{\encodingdefault}{\sfdefault}{m}{sl}
\SetMathAlphabet{\mathsfit}{bold}{\encodingdefault}{\sfdefault}{bx}{n}
\newcommand{\tens}[1]{\bm{\mathsfit{#1}}}

\def\tD{{\tens{D}}}

\def\tF{{\tens{F}}}

\def\tM{{\tens{M}}}

\def\tX{{\tens{X}}}

\usepackage{xspace}
\makeatletter
\DeclareRobustCommand\onedot{\futurelet\@let@token\@onedot}
\def\@onedot{\ifx\@let@token.\else.\null\fi\xspace}

\def\eg{\emph{e.g}\onedot}

\def\etc{\emph{etc}\onedot} 
\def\wrt{w.r.t\onedot} 

\makeatother

\icmltitlerunning{Taylor Videos for Action Recognition}

\begin{document}

\twocolumn[
\icmltitle{Taylor Videos for Action Recognition
}



\icmlsetsymbol{equal}{*}

\begin{icmlauthorlist}
\icmlauthor{Lei Wang}{equal,yyy}
\icmlauthor{Xiuyuan Yuan}{equal,yyy}
\icmlauthor{Tom Gedeon}{sch}
\icmlauthor{Liang Zheng}{yyy}
\end{icmlauthorlist}

\icmlaffiliation{yyy}{School of Computing, Australian National University, Canberra, Australia}
\icmlaffiliation{sch}{School of Electrical Engineering, Computing and Mathematical Sciences, Curtin University, Perth, Australia}

\icmlcorrespondingauthor{Lei Wang}{lei.w@anu.edu.au}

\icmlkeywords{Machine Learning, ICML}

\vskip 0.3in
]



\printAffiliationsAndNotice{\icmlEqualContribution} 

\begin{abstract}
Effectively extracting motions from video is a critical and long-standing problem for action recognition. 
This problem is very challenging because motions (i) do not have an explicit form, (ii) have various concepts such as displacement, velocity, and acceleration, and (iii) often contain noise caused by unstable pixels. Addressing these challenges, we propose the Taylor video, a new video format that highlights the dominant motions (\eg, a waving hand) in each of its frames named the Taylor frame. 
Taylor video is named after Taylor series, which approximates a function at a given point using important terms. In the scenario of videos, we define an implicit motion-extraction function which aims to extract motions from video temporal blocks. In these blocks, using the frames, the difference frames, and higher-order difference frames, we perform Taylor expansion to approximate this function at the starting frame. We show the summation of the higher-order terms in the Taylor series gives us dominant motion patterns, where static objects, small and unstable motions are removed. Experimentally, we show that Taylor videos are effective inputs to popular architectures including 2D CNNs, 3D CNNs, and transformers. When used individually, Taylor videos yield competitive action recognition accuracy compared to RGB videos and optical flow. When fused with RGB or optical flow videos, further accuracy improvement is achieved. Additionally, we apply Taylor video computation to human skeleton sequences, resulting in Taylor skeleton sequences that outperform the use of original skeletons for skeleton-based action recognition. Code is available at: \href{https://github.com/LeiWangR/video-ar}{https://github.com/LeiWangR/video-ar}.

\end{abstract}

\section{Introduction}



Extracting motions and thus recognizing actions from videos are important problems. 
Extensive efforts have been made in improving video inputs {\cite{kim2022capturing, Bilen_2016_CVPR, BilenTPARMI,wang2024flow}}, optimizing networks {\cite{i3d_net,wang2018temporal,wang2023videomaev2,lin2019tsm}}, and training skills {\cite{zeroshot,oneshot,fewshot}}. This paper focuses on improving the input, while leaving the rest unchanged. 

\begin{figure}[tbp]
    \centering
    \includegraphics[width=\linewidth]{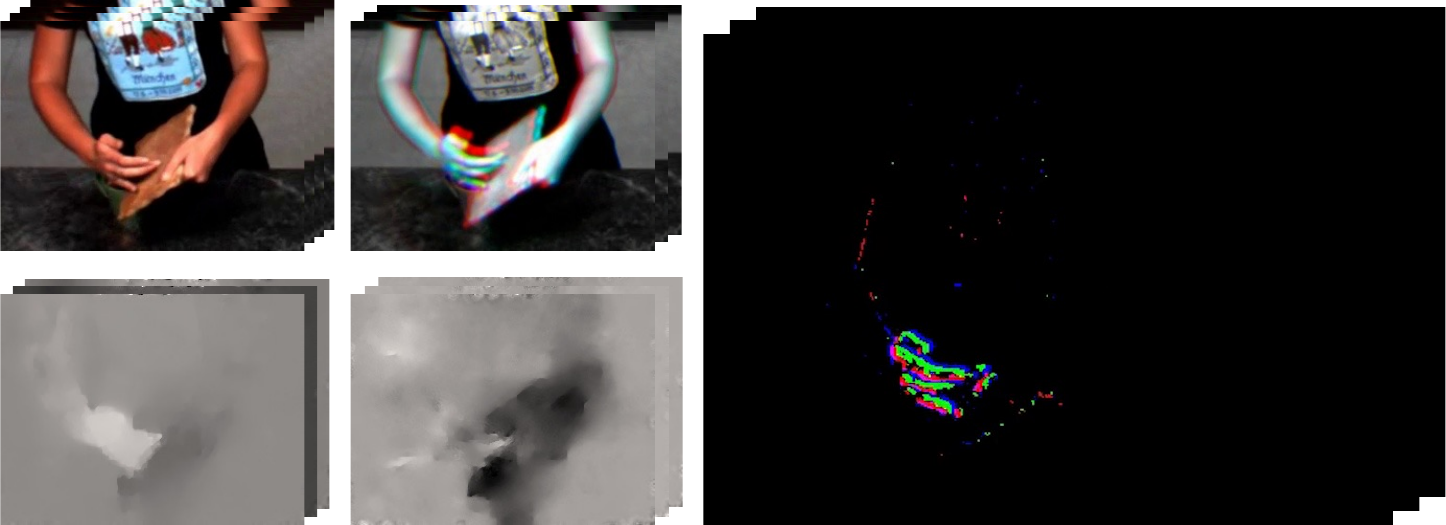}
    \caption{Visualizing different video formats. ({Top left}): RGB video and time-color reordering frames~\cite{kim2022capturing}. ({Bottom left}): $\mU$ and $\mV$ components of optical flow. ({Right}): proposed Taylor video frames.
    Taylor frames clearly (i) remove static objects and unstable motions and (ii) highlight motions. 
    }
    \label{fig:enter-label}
\end{figure}

Motion is an abstract and general concept. By `abstract', motion does not have an \textit{explicit} form, so we ask whether we can use a function to \textit{implicitly} capture motions in different granularity levels. In comparison, existing methods \textit{explicitly} define motions in different formats, such as optical flow that tracks pixels and objects.
By `general', there are various motion concepts, such as displacement, velocity, and acceleration. While they look challenging to compute, they have clear physical relationships, \eg, velocity is the derivative of displacement with respect to time. In action recognition, this has not been explicitly considered, to the best of our knowledge. 

We consider the above points and propose to model motion using an implicit function that outputs motion in some unknown format. Instead of seeking an exact function output which is intractable (motion is very abstract), we propose to approximate this output, especially the dominant motions, using video frames, their differences, and higher-order differences. We find that Taylor series is a great tool for this, allowing us to use similar approximation formulas to compute displacement, velocity, and acceleration.

\begin{figure*}[tbp]
    \centering
    \includegraphics[trim=0cm 2.8cm 0cm 1.8cm, clip=true, width=\textwidth]{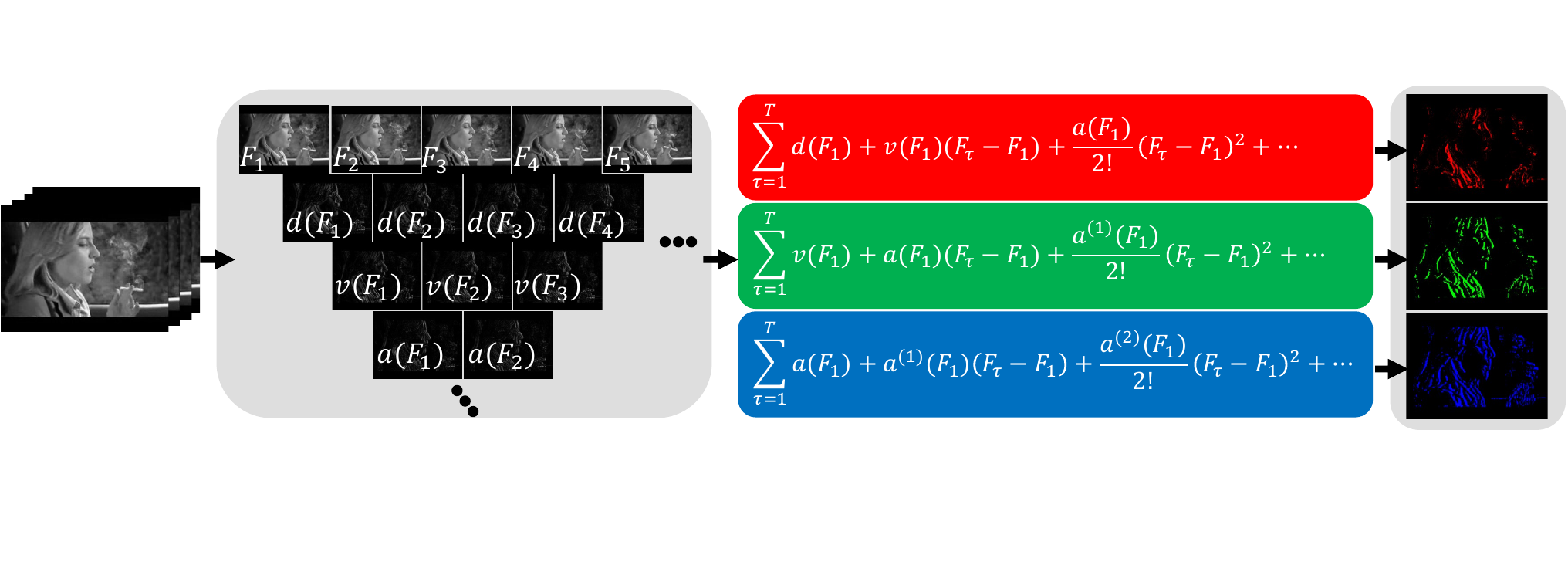}
    \caption{Computing a single Taylor frame from a grayscale video temporal block  
    $\tF\!=\![\mF_1, \mF_2, \cdots, \mF_\tau, \cdots,\mF_T], \tau\!=\!1, 2, \cdots, T$. We calculate the difference map between each two consecutive frames: $d(\mF_i)\!=\!\mF_{i\!+\!1}\!-\!\mF_i$, $i\!=\!1, 2, \cdots, T$. We then calculate the higher-order differences, \eg, velocity maps using $v(\mF_i)\!=\!d(\mF_{i\!+\!1})\!-\!d(\mF_i)$, acceleration maps using $a(\mF_i)\!=\!v(\mF_{i\!+\!1})\!-\!v(\mF_i)$, jerk maps, \etc, in the temporal block.
We compute three channels of a Taylor frame by Eq. \eqref{eq:displace},~\eqref{eq:velocity}, and~\eqref{eq:acc}, visualized in red, green, and blue, respectively.  
    }
    \label{fig:taylorframe}
\end{figure*}

In this paper, we propose a video format named Taylor video, which is composed of Taylor frames, to extract motions for human action recognition, as an alternative to RGB videos and optical flow. Fig.~\ref{fig:enter-label} shows a comparison. 
Taylor videos are directly converted from RGB videos.
Each Taylor frame has three channels, representing displacement, velocity, and acceleration of motion, respectively. 

To compute a Taylor frame from a temporal block, we first define an implicit motion-extraction function that uses the last frame as input and outputs the encoded motion. We then perform Taylor expansion of this implicit function, and sum up the important terms comprising of derivatives of the motion-extraction function. Here, derivatives are differences and higher-order differences of the frames in the temporal block.
By defining the motion-extraction functions to extract displacement, velocity, and acceleration, the difference frames are used as different orders in Taylor series, thus producing the displacement, velocity, and acceleration channels,  
comprising a Taylor frame.  Multiple Taylor frames constitute a Taylor video. Fig.~\ref{fig:taylorframe} shows an overview of forming a single Taylor frame.





Methodologically, Taylor videos have a few important advantages. First, compared with optical flow, it is much faster to compute due to its in-place matrix operations, and is not subject to computing errors due to its mathematical nature. Second, compared with RGB images, it gets rid of redundant static content, tiny and unimportant motions. Third, Taylor expansion allows for controllable motion capture: if fewer derivative terms are used, we capture dominant motions; if more terms are summed, more motion details are included. Lastly, Taylor videos benefit from the high resolution of RGB images while event cameras do not. 

Experimentally, Taylor videos are very competitive inputs compared with RGB and optical flow videos, when applied as the sole input. If combined, further improvements are observed, demonstrating their complementary nature. This is confirmed using various existing action recognition networks, including 2D CNNs, 3D CNNs, and transformers. 
The main points of this paper are summarized below. 
\renewcommand{\labelenumi}{\roman{enumi}.}
\begin{enumerate}[leftmargin=0.6cm]
    \item We introduce Taylor videos as an alternative to RGB videos and optical flow to extract motions for action recognition. Its computation comes from a decent application of Taylor series to videos.  
    
    \item Taylor videos are quick to compute from RGB videos, can be of high resolution, and can dynamically capture different levels of dominant motions. 
    
    \item We demonstrate Taylor videos are competitive with and also complementary to RGB videos and optical flow. 
    \item We apply Taylor video computation to human skeleton sequences, improving skeletal action recognition over original skeletons.
\end{enumerate}

\section{Related Work}

\textbf{Conventional videos and relevant networks.}
RGB videos are mostly commonly used, 
with each pixel having three color channels. 
They do not encode much temporal dynamics in their individual frames, necessitating architectures for temporal modeling and reasoning. Popular models for this purpose include IDT~\citep{dense_mot_boundary}, two-stream networks~\citep{two_stream}, 3D spatio-temporal features~\citep{spattemp_filters}, and spatio-temporal ResNet models~\citep{spat_temp_resnet}. Recent advanced models include \citep{bertasius2021space, qin2022fusing,liu2022video, radford2021learning, ni2022expanding, wu2022transferring,wang20233mformer,wang2023robust,wang2024high,wang2024meet}.

A few closely related works are \citep{wang2018temporal, wang2021tdn}. To reduce video redundancy, they compute differential images by subtracting pairs of consecutive frames, providing the network with a more explicit representation of high-frequency temporal changes. But they suffer from spatial shifting, such as camera jitter. In comparison, Taylor frames reduce those jittering motions and retain the dominant ones, and effectively improve action recognition performance.

 \textbf{Optical flow and relevant networks.} 
Optical flow is computed between consecutive video frames and a widely used secondary input for action recognition~\cite{two_stream, i3d_net, wang2024flow}. It contains both the direction and magnitude of motion at each pixel. Common methods for computing optical flow in action recognition include~\cite{tvl1_opt, ldof, deepflow, epicflow}. Because it is computationally expensive, recent works~\cite{zhang2016real, wu2018compressed} use motion vectors from compressed videos, \eg, H.264, to avoid high computational cost. 
In action recognition~\cite{two_stream, i3d_net}, optical flow is usually implemented in a separate stream, commonly referred to as the temporal stream, in addition to the appearance stream (\textit{a.k.a}. RGB stream). 

The Taylor video is very different from optical flow. While optical flow focuses more on pixel displacement\footnote{Its computation method is very different from ours.}, Taylor frames records more motion concepts including displacement, velocity and acceleration. Taylor frames can even account for higher order motion concepts such as jerk (derivative of acceleration), snap, crackle, \textit{etc}. Taylor videos are also much faster to compute. 





\textbf{Extracting temporal dynamics in a single frame.} \citet{Bilen_2016_CVPR} propose dynamic images 
that record spatial and temporal information in a single frame, allowing image classification networks to be used for action classification. Recently, \citet{kim2022capturing} introduced a channel sampling method that puts together R (G, or B) channels of consecutive frames into a single frame, allowing 2D CNNs to better capture motion.   
On the down side, these alternative video representations still contain lots of redundancy inherited from RGB videos. Moreover, they lack flexibility to extract various orders of motions. In comparison, the Taylor video is backed by well-founded mathematical concepts, reduces redundancy significantly, and is very flexible.




\section{Proposed Approach}
\label{sec:app}


\textbf{Notations.} 
Scalars are written in regular fonts; vectors are denoted by lowercase boldface letters, \eg, $\vx$; matrices by uppercase boldface, \eg, \boldsymbol{$\mX$}; tensors by calligraphic letters, \eg, $\tX$. Let $\tX\in\mbr{d_1\times d_2\times d_3}$ denote a third-order tensor. Using the Matlab convention, we refer to its $k$-th slice as $\tX_{:,:,k}$, which is a $d_1\!\times\!d_2$ matrix.

\textbf{Taylor series revisit.} 
Taylor series locally approximates non-linear functions. It is an infinite sum of terms expressed in terms of the function's derivatives at a single point: 
\begin{equation}
    f(x)=\sum_{k=0}^\infty \frac{f^{(k)}(a)}{k!}(x-a)^k,
    \label{eq:taylorseries}
\end{equation}
where $k!$ denotes the factorial of $k$. $f^{(k)}(a)$ denotes the $k$th derivative of $f$ evaluated at point $a$. The 0th-order derivative of $f$ is defined as $f$ itself, and $(x-a)^0$ and $0!$ are both equal to 1. Theoretically, the first few terms of the series can reconstruct most of 
$f(x)$. This forms the foundation of our use: the first few terms encode dominant motions.

\subsection{Taylor videos: General formulations}

\textbf{Taylor series to compute motion within a temporal block.} For an RGB video, we first convert each frame to grayscale. 
We are given a $\mathcal{T}$-frame grayscale video $\tF\!=\![\mF_1, \mF_2, \cdots, \mF_\mathcal{T}] \!\in\!\mbr{H\!\times\!W\!\times\!\mathcal{T}}$, where $H$ and $W$ denote respectively the height and width. 
We use a temporal sliding window sized $T$ with step size 1 to produce individual subsequences (\textit{a.k.a.} video temporal blocks), resulting in $N$ temporal blocks denoted as $\{\tF^1, \tF^2, \cdots, \tF^N\}$, where $\tF^i\!=\![\mF^i_1, \mF^i_2, \cdots, \mF^i_{T}]\!\in\!\mbr{H\!\times\!W\!\times\!T}$. For simplicity we drop superscript $i$ in what follows unless otherwise stated.

We define a motion extraction function $f: \mF \in \mbr{H\!\times\!W} \mapsto \mM \in \mbr{H\!\times\!W}$, where the function input argument $\mF$ is a gray-scale frame, 
and the output $\mM$, presenting pixel-level motion, is the motion map encoded in the temporal block ending at $\mF$. 
%
Given a temporal block, we aim to reveal its motion dynamics. To achieve this, we propose to use Taylor series with Eq.~\eqref{eq:taylorseries}:
%
\begin{equation}
    f(\mF_{T})=\sum_{k=0}^\infty \frac{f^{(k)}(\mF_1)}{k!}\odot(\mF_{T}-\mF_1)^{\circ k}, 
    \label{eq:talor}
\end{equation}
%
where $\odot$ and $^{\circ k}$ denote the Hadamard (element-wise) product and Hadamard (element-wise) power, respectively. $f^{(k)}(\mF_1)$ denotes the $k$th derivative of $f$ evaluated at $\mF_1$. 


\begin{figure*}[tbp]
    \centering
    \includegraphics[width=\textwidth]{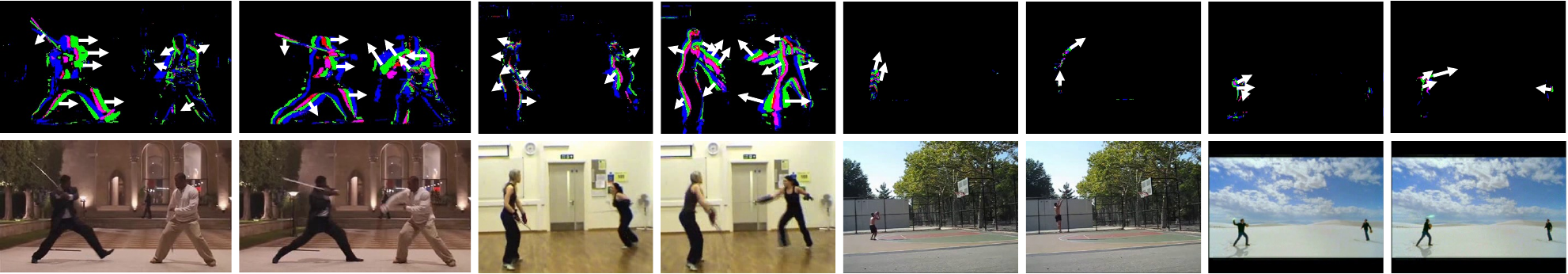}
    \caption{Taylor frames indicate motion strengths and directions. (Top): Taylor frames. (Bottom): original RGB frames. All videos are from HMDB-51. 
    Red, green, and blue represent displacement, velocity, and acceleration, respectively. Bolder colors indicate greater strength. We depict motion directions with white arrows: if green (velocity) is to the right of red (displacement), the object is moving rightwards \textit{in the next frame}; if blue (acceleration) is to the left of red (displacement), the object is moving leftwards \textit{in the frame after}.} 
    \label{fig:strength}
\end{figure*}

\textbf{Combining short-term and long-term motions in a temporal block.} Eq. \eqref{eq:talor} computes motions covering the entire temporal block with length $T$, seen through the term $\mF_{T}-\mF_1$. In order to consider shorter-term motions in a temporal block, we also compute motions between frame $\mF_\tau$ ($\tau\!<\!T$) and frame $\mF_1$ using Eq. \eqref{eq:talor},  
and then average them. 
This process can be described as:
\begin{equation}
    \mM_{f} = \frac{1}{T}\sum_{\tau=1}^{T}f(\mF_{\tau}).
    \label{eq:sum-motion-concept}
\end{equation}
where $f(\mF_{\tau})=\sum_{k=0}^\infty \frac{f^{(k)}(\mF_1)}{k!}\odot(\mF_{\tau}-\mF_1)^{\circ k}$, and $\tau=1, 2, \cdots, T$.
%
%
%
%
%
Here, 
%
$\mM_{f}$ represents the motions encoded in the temporal block ending with $\mF_{\tau}$. Here subscript $f$ is used to denote extracting a certain motion concept, which can be one of displacement, velocity, and acceleration.

\subsection{Computing three motion concepts in Taylor videos}
\label{sec:motion-concepts}

By defining different motion extraction functions, 
we use Eq \eqref{eq:sum-motion-concept} to compute three critical motion concepts: displacement, velocity, and acceleration within a temporal block.  
%

\textbf{Displacement} refers to the change in position of a pixel, and it has both magnitude and direction. 
To approximate displacement, we define displacement-extraction function $f_d$, which takes a frame as input and outputs displacement within a temporal block. After aggregating the short-term and long-term motions using Eq. \eqref{eq:sum-motion-concept}, the displacement motion map can be approximated as: 
\begin{equation}
    \mM_{d} = \frac{1}{T}\sum_{\tau=1}^{T}
    \sum_{k=0}^\infty \frac{f_d^{(k)}(\mF_1)}{k!}\odot(\mF_{\tau}-\mF_1)^{\circ k}.
    \label{eq:displace}
\end{equation}
Here $f^{(0)}_d\!=\!d(\mF_1)\!=\!\mF_2\!-\!\mF_1$, $f^{(1)}_d\!=\!v(\mF_1)\!=\!d(\mF_2)\!-\!d(\mF_1)$, and $f^{(2)}_d\!=\!a(\mF_1)\!=\!v(\mF_2)\!-\!v(\mF_1)$ (assume the time step is 1). We take $v(\mF_1)$ as an example to illustrate the intuition behind these equations. Because $f_d$ characterizes the displacement of the temporal block, its first-order derivative, or $\frac{d_{f_d}}{d_\mF}$ means the displacement change between two consecutive frames, which should be intuitively computed as the second-order difference between two gray-scale frames.

%

\textbf{Velocity} describes the rate of change of displacement, and includes both speed and direction. Similar to the computation of displacement,  
we form velocity motion map as:
\begin{equation}
\mM_{v} = \frac{1}{T}\sum_{\tau=1}^{T}
    \sum_{k=0}^\infty \frac{f_v^{(k)}(\mF_1)}{k!}\odot(\mF_{\tau}-\mF_1)^{\circ k}.
    \label{eq:velocity}
\end{equation}
Similarly, $f^{(0)}_v\!=\!v(\mF_1)\!=\!d(\mF_2)\!-\!d(\mF_1)$, $f^{(1)}_v\!=\!a(\mF_1)\!=\!v(\mF_2)\!-\!v(\mF_1)$, and $f^{(2)}_v\!=\!j(\mF_1)\!=\!a(\mF_2)\!-\!a(\mF_1)$. Jerk $j(\mF_1)$ is the rate of change of acceleration at frame $\mF_1$.

%
\textbf{Acceleration} describes the rate of change of velocity. 
%
Similar to displacement and velocity, the acceleration map of a given temporal block can be approximated as:
\begin{equation}
\mM_{a} = \frac{1}{T}\sum_{\tau=1}^{T}
    \sum_{k=0}^\infty \frac{f_a^{(k)}(\mF_1)}{k!}\odot(\mF_{\tau}-\mF_1)^{\circ k}.
    \label{eq:acc}
\end{equation}
%
%
Here $f^{(0)}_a\!=\!a(\mF_1)\!=\!v(\mF_2)\!-\!v(\mF_1)$, and $f^{(1)}_a\!=\!j(\mF_1)\!=\!a(\mF_2)\!-\!a(\mF_1)$.
Given a temporal block, we compute three motion maps $\mM_{d}$, $\mM_{v}$, and $\mM_{a}$. These maps are 
stacked into an image, forming a \textbf{Taylor frame}:  $\mM\!\in\!\mbr{H\!\times\!W\!\times\!3}$. Given an RGB video that has $N$ temperal blocks: $\{\tF^1, \tF^2, \cdots, \tF^N\}$, we compute $N$ Taylor frames $\mM^1, \mM^2, \cdots, \mM^N$. These Taylor frames form the \textbf{Taylor video}: $\tM\!=\![\mM^1, \mM^2, \cdots, \mM^N]\!\in\!\mbr{H\times W \times 3 \times N}$. Fig.~\ref{fig:taylorframe} illustrates the computation of a single Taylor frame. In Fig. \ref{fig:strength}, we show that Taylor frames allow us to visually perceive the strength and direction of motion.


%
%

\subsection{Efficient computation of Taylor videos}


Given a temporal block $\tF\!\in\!\mbr{H\!\times\!W\!\times\!T}$ in the form of a tensor, we create an inflated tensor (mimicking a static temporal block) $\widetilde{\tF}\!\in\!\mbr{H\!\times\!W\!\times\!T}$ by duplicating the first frame of the temporal block $T$ times. 
Instead of using $f$ that uses a frame as input, we use function $t$ that takes a video temporal block, or a tensor as input, and outputs a single three-channel Taylor frame that summarises the motions. The proposed tensor representation is given below:
\begin{equation}
    t(\tF)=\sum_{k=0}^\infty \frac{t^{(k)}(\widetilde{\tF})}{k!}\odot\frac{1}{T}\sum_{\tau=1}^{T}(\tF-\widetilde{\tF})^{\circ k}_{:,:,\tau}.
    \label{eq:tensor}
\end{equation}
%
%
%
%
Eq. \eqref{eq:tensor} is equivalent to Eq. \eqref{eq:sum-motion-concept} giving the same Taylor frame output, and is more efficient to compute because of the use of tensors. A proof of the equivalence is presented in Section~\ref{sec:appendix-proof} of Appendix. 
Algorithm~\ref{alg:taylor} in Section~\ref{sec:alg} of Appendix shows the efficient implementation of Taylor video. 

\begin{figure*}[tbp]
    \centering
    \includegraphics[width=\linewidth]{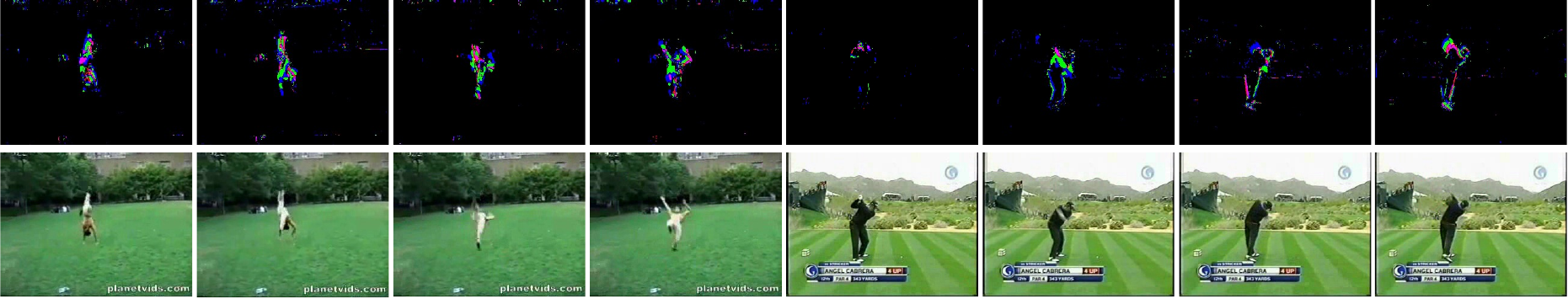}
    \caption{
    Taylor videos remove redundancy, such as static backgrounds, unstable pixels, watermarks, and captions. This, together with its ability to highlight motion including strengths and directions, is beneficial for action recognition. Videos are from HMDB-51. }  
    \label{fig:background}
\end{figure*}

\subsection{Discussion}

\textbf{Can we sum up an infinite number of terms in Eq.~\eqref{eq:talor}?} No. An infinite number of terms require an infinite number of frames in a temporal block, which is infeasible. 

\textbf{Can we compute more motion concepts in Taylor video?} Technically yes, by extending Eq.~\eqref{eq:sum-motion-concept} into jerk, \etc. However, doing this has two down sides. First, computing higher-order motions requires heavier computation, because we need to compute more frame differencing operations. Second, if we include more channels ($>\!3$) for Taylor videos, we will have to modify model architecture, and it is non-trivial when using pre-trained models. In fact, Taylor videos with three channels can easily replace RGB video inputs and reuse existing pre-trained models. Taylor videos have superior performance on pre-trained models compared with being trained from scratch.

\textbf{How to choose the number of frames in a temporal block and the number of terms in Taylor series?} Because of the discrete nature of frames, and the fact that higher-order image differences are fewer than low-order ones, we need at least 4 frames per temporal block to compute a 3-channel Taylor frame. Any temporal block with fewer than 4 terms will result in insufficient data for Taylor frame computation. If a temporal block has $T\!>\!4$ frames, we recommend to use $T\!-\!3$ terms to fully exploit data in the block. If we use fewer than $T\!-\!3$ terms, some frames will not be used; but it is our future work to explore how to sample frames so that we can use few terms to reduce computation cost. Empirical analysis of the impact of $T$ is provided in Section \ref{sec:eval}.

\textbf{Connection with the event camera.}
Event cameras continuously capture differences in pixel brightness. From a visualization standpoint, the displacement channel $M_d$ of the Taylor video appears similar to event camera data. However, utilizing event camera data for action recognition necessitates distinct temporal preprocessing due to its high temporal resolution in capturing fine-grained temporal dynamics. This characteristic differs from traditional frame-based cameras, which may overlook rapid changes.
Taylor videos are computed from conventional videos, the most popular video format, focusing on extracting dominant motions using frame differencing as the measure of time concept, \eg, $\mF_{\tau}-\mF_1$, $\mF_{\tau\!-\!1}-\mF_1$, \etc, instead of using discrete time steps, \eg, $\tau\!-\!1$, $\tau\!-\!2$, \etc.

\textbf{Taylor skeleton sequences.} Note that Taylor videos can be naturally applied to human skeleton sequences. Similar to the RGB modality, we extract dominant motions for each human body joint within the video temporal block. By applying Taylor series to skeleton sequences, we obtain Taylor skeleton sequences. 

\textbf{Limitation}. Taylor video obviously does not encode much static texture pattern. It means Taylor videos at their current form are not an ideal option for tasks like video captioning and anomaly detection. A potential solution is to stack some RGB / grayscale frames to Taylor frames so that static objects and backgrounds can be considered for more general video understanding, such as large video language models. However, the Taylor video effectively removes backgrounds, watermarks, captions, \etc, while retaining significant motions (refer to samples from HMDB-51 in Fig.~\ref{fig:background}).

\textbf{Other potential applications}. Because Taylor videos capture dominant motions, it may be used in tasks like video-based face anti-spoofing. Taylor videos also present motion strength and direction\footnote{Each channel of the Taylor frame represents a motion concept with positive and negative values indicating motion directions (0 for static pixels). Velocity and acceleration channels are computed per video temporal block, capturing relative motion directions from the initial frame.} (see Fig.~\ref{fig:strength}), they can potentially be used for player performance analysis.   
Moreover, Taylor videos are valuable in video analysis for predicting the next frames or future motion trajectories, aiding tasks such as object tracking and scene understanding, as they encode both short-term and long-term motions per Taylor frame.

\section{Experiments}\label{sec:exp}

\begin{table*}
	\begin{minipage}{0.67\linewidth}
		\centering
  \setlength{\tabcolsep}{4.0pt}
  \begin{tabular}{l l l l c c c c}
\toprule
 & \multirow{2}{*}{Model} & \multirow{2}{*}{Pretrain} & \multirow{2}{*}{Input}   & \multirow{2}{*}{HMDB-51} & \multicolumn{2}{c}{CATER} & \multirow{2}{*}{MPII}\\
\cline{6-7}
& & & && {static} & {moving}  &  \\
\midrule
\multirow{5}{*}{\rotatebox[origin=c]{90}{\bf 2D CNNs}} & \multirow{2}{*}{TSN} 
& \multirow{2}{*}{ImageNet} & RGB & 54.9 &  49.6 & 51.6 & 38.4\\
&  & & \cellcolor{gray!10}Taylor & \cellcolor{gray!10}56.4 & \cellcolor{gray!10}73.8 & \cellcolor{gray!10}62.7&\cellcolor{gray!10}42.2\\
\cdashline{2-8}
& \multirow{3}{*}{TSM} & \multirow{3}{*}{ImageNet} & RGB & - & 79.9 & 65.8 & 46.7 \\

&   & & GrayST & - & 82.2 & 74.7 & 48.7\\
&   & & \cellcolor{gray!10}Taylor & \cellcolor{gray!10}- & \cellcolor{gray!10}83.1 & \cellcolor{gray!10}75.5 & \cellcolor{gray!10}50.1\\
\hline
\multirow{7}{*}{\rotatebox[origin=c]{90}{\bf 3D CNNs}}
& \multirow{5}{*}{I3D} & \multirow{2}{*}{ImageNet} & RGB   & 49.8 & 73.5  & 57.7&  42.8\\
&   & & \cellcolor{gray!10}Taylor  & \cellcolor{gray!10}65.2 & \cellcolor{gray!10}74.7 & \cellcolor{gray!10}60.5 & \cellcolor{gray!10}43.0 \\ 
\cdashline{3-8}
&  &\multirow{3}{*}{Kinetics} & RGB   & 74.3 & 75.4 & 61.9 & 48.7 \\
& & & OPT &  77.3 & 78.5  & 66.3 &  51.0\\
&   & & \cellcolor{gray!10}Taylor  & \cellcolor{gray!10}78.1 & \cellcolor{gray!10}80.2 & \cellcolor{gray!10}69.8&  \cellcolor{gray!10}52.3\\ 
\cdashline{2-8}
& \multirow{2}{*}{R(2+1)D} &\multirow{2}{*}{Sports1M} & RGB   & 66.6 & - & -&-\\
&  & & \cellcolor{gray!10}Taylor  & \cellcolor{gray!10}67.4 & \cellcolor{gray!10}- & \cellcolor{gray!10}-& \cellcolor{gray!10}- \\ 
\hline
\multirow{4}{*}{\rotatebox[origin=c]{90}{\bf Transf.}}
& \multirow{2}{*}{TimeSformer} &\multirow{2}{*}{Kinetics} & RGB  & 71.7 & 69.9 & 57.6 &  41.0 \\
&  & & \cellcolor{gray!10}Taylor & \cellcolor{gray!10}72.1& \cellcolor{gray!10}71.2 & \cellcolor{gray!10}58.2&  \cellcolor{gray!10}42.8\\
\cdashline{2-8}
& \multirow{2}{*}{Swin Transformer} &\multirow{2}{*}{Kinetics} & RGB  & 72.9 & 72.2 & 63.5&  46.6\\

& & & \cellcolor{gray!10}Taylor & \cellcolor{gray!10}73.5 & \cellcolor{gray!10}73.0 & \cellcolor{gray!10}64.7 & \cellcolor{gray!10}47.0\\
\bottomrule
\end{tabular}
\caption{Comparing the Taylor video with other input modalities on three datasets with various action recognition models and pre-training datesets.}
\label{tab:globaltable}
	\end{minipage}\hfill
	\begin{minipage}{0.30\linewidth}
		\centering
\includegraphics[width=0.85\linewidth]{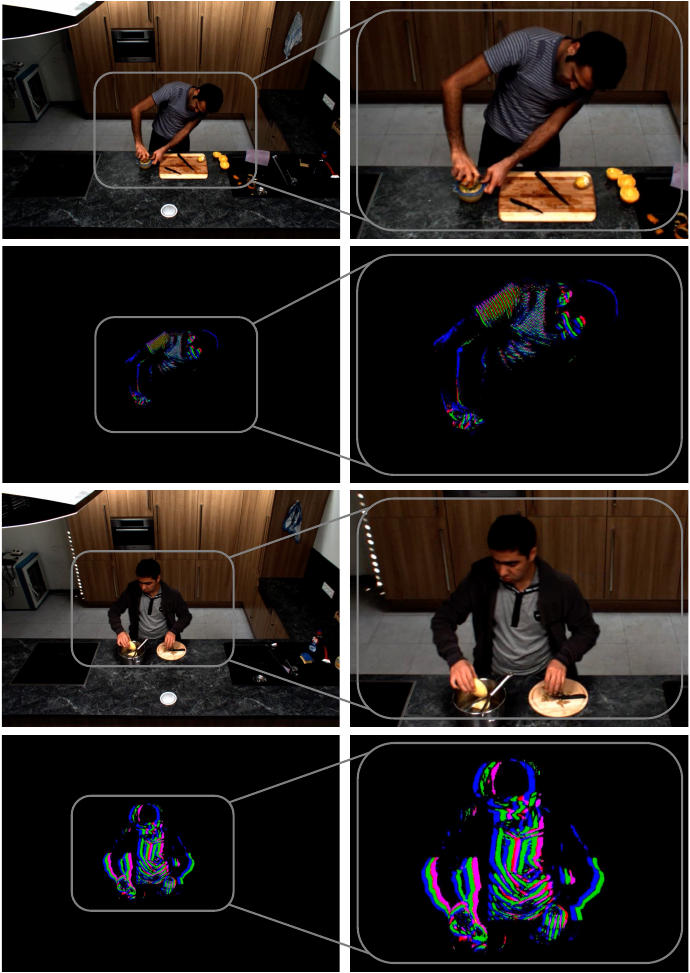}
    \captionof{figure}{
    Taylor frame captures subtle motions on MPII. (\textit{Top 4 images}) show \textit{squeeze} and (\textit{Bottom 4 images}) show \textit{put in pan/pot}. In each set, the left motion region is zoomed in on the right to enhance visualization. Better view in color.
    }
  \label{fig:finegrained}
	\end{minipage}
\end{table*}

\begin{table}[!tbp]
    \centering
    \setlength{\tabcolsep}{4.0pt}
    \begin{tabular}{lccccc}
        \toprule
         & TSN & TSM &  C3D & I3D & R(2+1)D\\
        \midrule
        RGB & 49.6 & 40.3 & 24.3 & 43.9 & 26.0\\
        Taylor & 49.8 & 41.3 & 25.2 & 44.4 & 26.9\\
        \bottomrule
    \end{tabular}
    \caption{Comparing RGB videos and Taylor videos under the training-from-scratch setup. Various action recognition models are used on HMDB-51. Top-1 accuracy (\%) is reported.}
    \label{tab:scratch}
\end{table}

\subsection{Datasets and implementation details}

\textbf{Datasets}. First, we use three small-scale datasets: HMDB-51~\cite{kuehne2011hmdb}, MPII Cooking Activities~\cite{rohrbach2012database}, and CATER~\cite{girdhar2020cater}. HMDB-51 has 51 human action categories and 6766 videos. 
It is challenging due to its significant camera and background motion. 
The MPII dataset has 64 distinct activities from 3748 clips include coarse actions such as \textit{opening refrigerator}, and fine-grained actions such as \textit{peel}, \textit{slice}, and \textit{cut apart}. 
While its cameras are fixed, the human actions usually occupy very small areas with relatively subtle movements.
CATER is a synthetic action recognition dataset involving long-term temporal reasoning. This provides the biggest challenge for action recognition methods that only focus on short-term clips. It has two versions of the videos, with and without camera motion, and we experiment with both of them.
%
Because of the diverse aspects focused by the datasets, we can evaluate Taylor videos \wrt its effectiveness for static and moving cameras, complex backgrounds, subtle human actions, and non-human movements. 
We then evaluate Taylor videos on large-scale Kinetics (K400 / K600)~\cite{kay2017kinetics} and Something-Something v2 (SSv2)~\cite{mahdisoltani2018effectiveness}. 
We also evaluate the effectiveness of Taylor skeleton sequences using NTU-60~\cite{Shahroudy_2016_NTURGBD}, NTU-120~\cite{Liu_2019_NTURGBD120}, and Kinetics-Skeleton (K-Skel)~\cite{stgcn2018aaai}.

\textbf{Evaluation protocols.} 
For HMDB-51, we use the standard 3 train/test splits and report the mean accuracy across 3 splits. For MPII, we use the mean Average Precision (mAP) over 7-fold cross validation. For CATER, we also report the mAP on both static and moving camera setups. For large-scale video and skeleton datasets, we adhere to standard evaluation protocols and report Top-1 accuracy.

\textbf{Implementation details.} 
Using Taylor videos as input, we either train action recognition models from scratch or fine-tune them on top of models pretrained with RGB image or videos. The model architectures cover popular 2D and 3D CNNs, such as TSM~\cite{lin2019tsm}, TSN~\cite{wang2018temporal}, TSN~\cite{wang2018temporal}, SlowFast~\cite{feichtenhofer2019slowfast}, R(2+1)D~\cite{tran2018closer}, C3D~\cite{spattemp_filters}, and I3D~\cite{i3d_net}, as well as recent transformer-based models such as Swin Transformer~\cite{liu2021swin} and TimeSformer~\cite{bertasius2021space}. We use pre-computed optical flow provided by the dataset websites. 
When implementing Taylor videos for transformer architectures, we add the grayscale frame to each of the displacement, velocity, and acceleration maps. Without this strategy, some patch embeddings of transformers would only be computed on patches that are almost all zeros, compromising transformer performance\footnote{In fact, existing transformer architectures are predominately designed for RGB videos. Adapting them to Taylor videos is an interesting future direction.}. 
We follow the default settings from the original papers, where we replicate their performance using RGB videos and/or optical flow as input. Hyperparameters such as the number of epochs for training/fine-tuning are determined on the validation sets. 

We present train-from-scratch results using either RGB or Taylor videos under the TSM, I3D, TimeSformer (L), Swin Transformer, and VideoMAE v2~\cite{wang2023videomae} backbones on large-scale video datasets. We also present experiments using ST-GCN~\cite{stgcn2018aaai}, InfoGCN~\cite{9879266}, AGE-Ens~\cite{qin2022fusing}, and 3Mformer~\cite{wang20233mformer} backbones on both original and Taylor skeleton sequences. We simply use the displacement concept with 1 term (4 frames per temporal block with a step size of 1) to compute the Taylor skeleton sequences.

\begin{table}[!tbp]
    \centering
    \resizebox{\linewidth}{!}{
    \setlength{\tabcolsep}{7.0pt}
    \begin{tabular}{llccc}
        \toprule
        Model & Input & K400 & K600 & SSv2\\
        \midrule
        \multirow{2}{*}{TSM} & RGB & 76.3 & - & 63.4 \\
        & \cellcolor{gray!10}Taylor & \cellcolor{gray!10}77.6 & \cellcolor{gray!10}- & \cellcolor{gray!10}65.1 \\
        \hline
        \multirow{2}{*}{I3D} & RGB & 77.7 & - & -\\
        & \cellcolor{gray!10}Taylor & \cellcolor{gray!10}79.3 & \cellcolor{gray!10}- & \cellcolor{gray!10}-\\
        \hline
        \multirow{2}{*}{TimeSformer} & RGB & 80.7 & 82.2 & 62.5 \\
        & \cellcolor{gray!10}Taylor & \cellcolor{gray!10}81.5 & \cellcolor{gray!10}83.1 & \cellcolor{gray!10}63.7 \\
        \hline
        \multirow{2}{*}{VideoMAE} & RGB & 79.8 & - & 69.3 \\
        & \cellcolor{gray!10}Taylor & \cellcolor{gray!10}80.4 & \cellcolor{gray!10}- & \cellcolor{gray!10}70.0 \\
        \hline
        \multirow{2}{*}{Swin Transformer} & RGB & - & - & 69.6\\
        & \cellcolor{gray!10}Taylor & \cellcolor{gray!10}- & \cellcolor{gray!10}- & \cellcolor{gray!10}71.1\\
        \bottomrule
    \end{tabular}}
    \caption{Evaluations of Taylor videos on large-scale Kinetics (K400 / K600) and Something-Something v2 (SSv2).}
    \label{tab:large-scale-taylor}
\end{table}

\begin{table}[t!]
\begin{center}
\resizebox{\linewidth}{!}{
\setlength{\tabcolsep}{2.0pt}
\begin{tabular}{l l c c c c c}
\toprule
\multirow{2}{*}{Model} & \multirow{2}{*}{Input} & \multicolumn{2}{c}{NTU-60} & \multicolumn{2}{c}{NTU-120} & K-Skel\\
\cline{3-7}
& & {X-Sub} & {X-View} & {X-Sub} & {X-Set} & Top-1\\
\midrule
\multirow{2}{*}{ST-GCN} & Skeleton & 81.5 & 88.3 & 70.7 & 73.2 & 30.7\\
& \cellcolor{gray!10}Taylor & \cellcolor{gray!10}85.4 & \cellcolor{gray!10}93.0 & \cellcolor{gray!10}78.5 & \cellcolor{gray!10}80.1 & \cellcolor{gray!10}35.1\\
\hline
\multirow{2}{*}{InfoGCN}& Skeleton & 93.0 & 97.1 & 89.8 & 91.2 & - \\
& \cellcolor{gray!10}Taylor & \cellcolor{gray!10}94.6 & \cellcolor{gray!10}98.5 & \cellcolor{gray!10}91.6 & \cellcolor{gray!10}93.7 &\cellcolor{gray!10} - \\
\hline
\multirow{2}{*}{AGE-Ens}& Skeleton & 91.0 & 96.1 & 87.6 & 88.8 & - \\
& \cellcolor{gray!10}Taylor & \cellcolor{gray!10}95.0 & \cellcolor{gray!10}98.3 & \cellcolor{gray!10}91.8 & \cellcolor{gray!10}92.5 &\cellcolor{gray!10} - \\
\hline
\multirow{2}{*}{3Mformer}& Skeleton & 94.8 & 98.7 & 92.0 & 93.8 & 48.3 \\
& \cellcolor{gray!10}Taylor & \cellcolor{gray!10}95.3 & \cellcolor{gray!10}98.8 & \cellcolor{gray!10}92.6 & \cellcolor{gray!10}94.7 & \cellcolor{gray!10}49.2 \\
\bottomrule
\end{tabular}}
\caption{Comparing Taylor-transformed skeletons with original skeletons on NTU-60, NTU-120 and Kinetics-Skeleton (K-Skel).}
\label{tab:taylor-skeleton}
\end{center}
\end{table}

\begin{figure*}[tbp]
    		\centering
\subfigure[I3D]{\label{fig:c}\includegraphics[trim=0cm 1.4cm 0cm 0cm, clip=true, height=21mm]{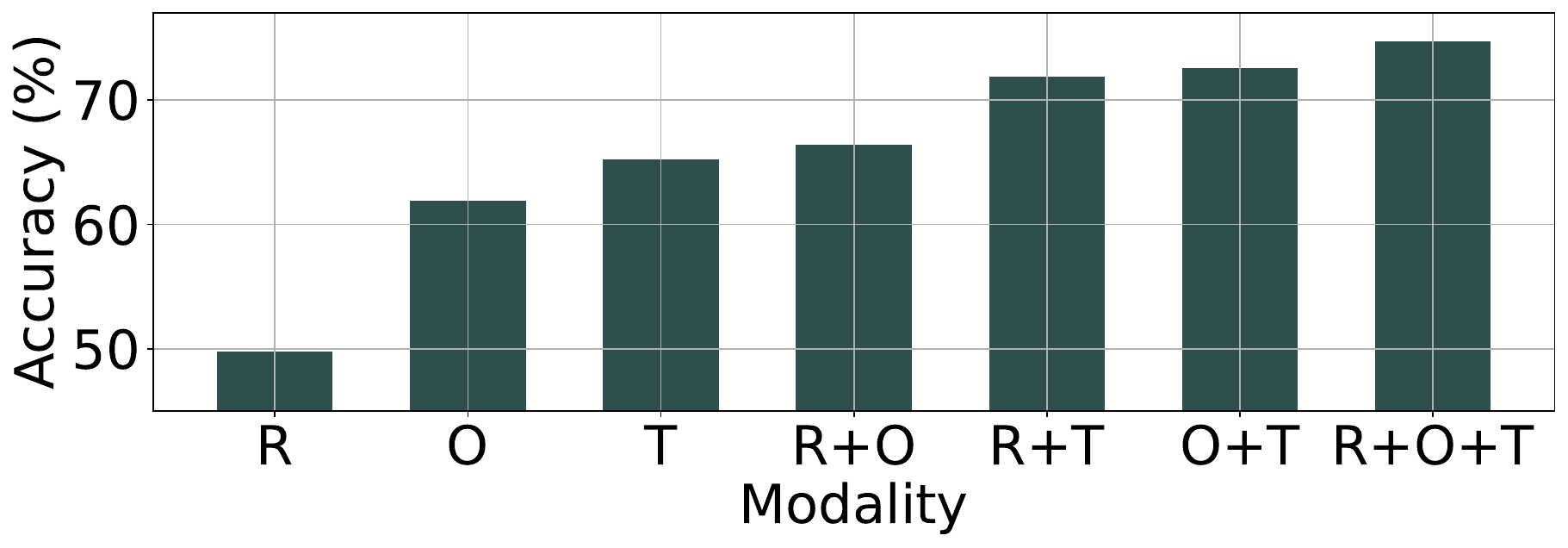}}\hfill
\subfigure[TimeSformer]{\label{fig:d}\includegraphics[trim=0cm 1.4cm 0cm 0cm, clip=true, height=21mm]{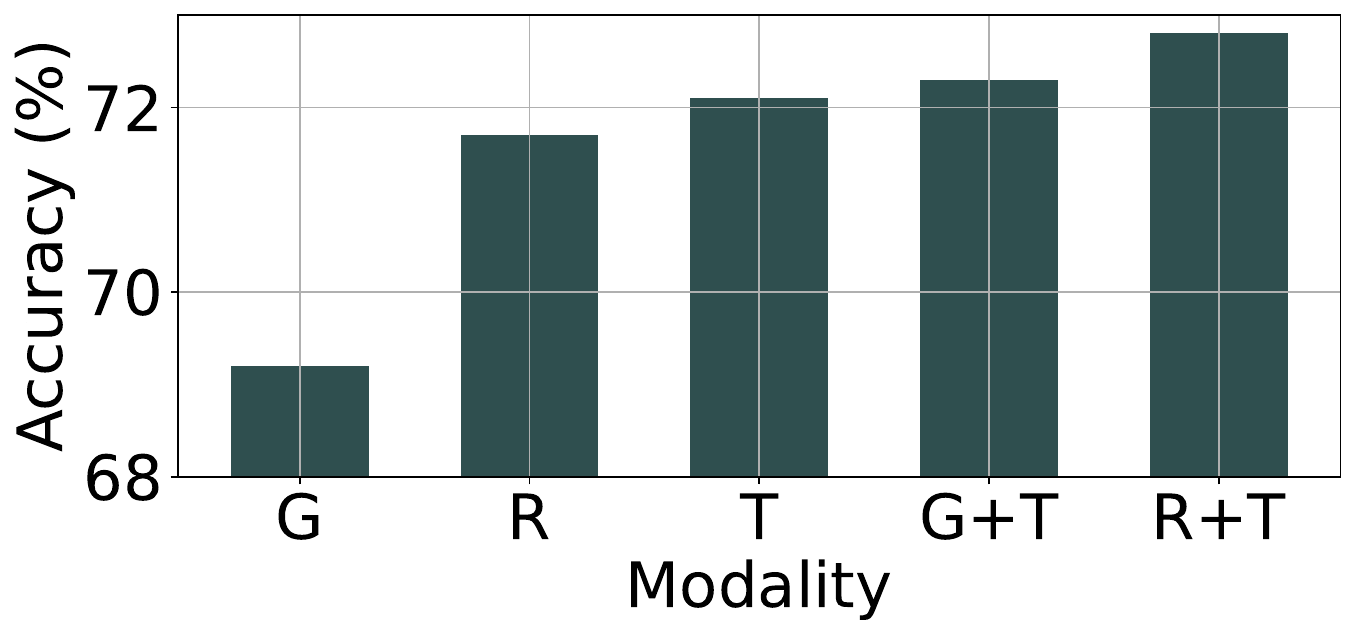}}\hfill
\subfigure[Impact of $T$]{\label{fig:termsvsacc}\includegraphics[trim=0cm 1.4cm 0cm 0cm, clip=true, height=21mm]{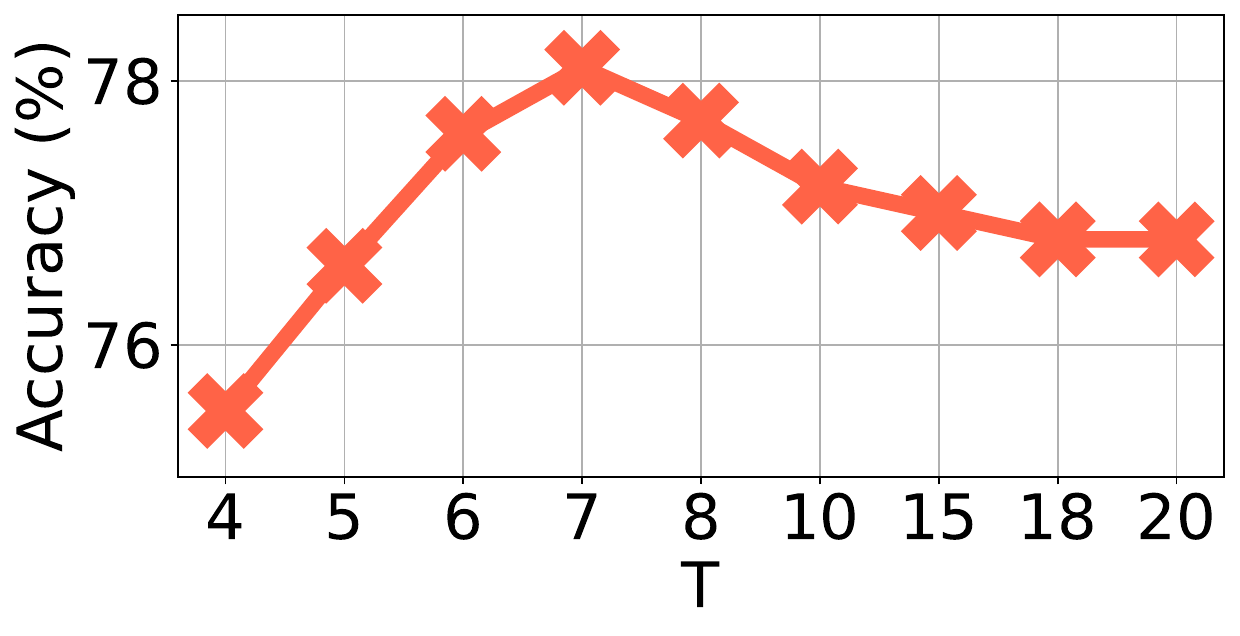}}
    \captionof{figure}{Comparison of different input modalities and their combinations using the (a) I3D and (b) TimeSformer models on HMDB-51. `R', `O' `G', and `T' denote RGB, optical flow, gray-scale, and Taylor videos, respectively. (c) Impact of the length $T$ of video temporal blocks for computing Taylor videos on the HMDB-51 dataset. Top-1 accuracy is reported.} 
  \label{fig:complement}
\end{figure*}

\subsection{Main evaluation}

\textbf{Comparing Taylor videos with RGB videos and optical flow}. In this experiment, we use only one modality as network input and compare the results. Various networks are used, including 2D CNNs, 3D CNNs, and transformers. We always use models pre-trained on various datasets before fine-tuning with the compared inputs. Results are summarized in Table \ref{tab:globaltable}. We have two main observations. 

First, on all the three datasets, using Taylor videos as input is very competitive compared with RGB videos, GrayST~\cite{kim2022capturing}, and optical flow. For example, under the I3D model with Kinetics pre-training, we achieve top-1 accuracy of 78.1\%, 80.2\%, 69.8\%, and 52.3\% on HMDB-51, CATER-static, CATER-moving, and MPII, respectively. Compared with RGB videos, the improvements are +3.8\%, +4.8\%, +7.9\%, and +3.6\% on the four test setups, respectively; compared with optical flow, the improvements are +0.8\%, +1.7\%, +3.5\%, and +1.3\%, respectively. 

Second, the competitiveness of Taylor videos can be observed under various models. For example, on the CATER-moving dataset, if we compare Taylor videos with RGB videos, the improvements are +11.1\%, +9.7\%, +2.8\%, +0.6\%, +1.2\% under TSN, TSM, I3D (ImageNet pretrained), TimeSformer, Swin-T models, respectively. This demonstrates the model-generic use of Taylor videos. 

These results indicate that Taylor video can deal with complex backgrounds with moving/static cameras (HMDB-51), long-term temporal reasoning (CATER), and fine-grained motions that only take up a small area (MPII, see Fig.~\ref{fig:finegrained}).

\textbf{Taylor video complements RGB videos and optical flow}. We now evaluate different combinations of these input modalities, including RGB, RGB + Taylor, RGB + optical flow, and RGB + optical flow + Taylor. Results are shown in Fig. \ref{fig:c} and~\ref{fig:d}. We have the following observations. 

First, under the I3D model, combining  Taylor with RGB or optical flow yields higher accuracy compared with using them individually. For example, using RGB+Taylor produces +22.1\% and +6.7\% improvement over using only RGB or Taylor, respectively. If we combine all the three modalities, further improvement is observed. 

Second, under TimeSformer, we show performance of Taylor videos with either RGB (R) or gray-scale (G) frames, and such combinations are again superior to individual modalities. On the HMDB-51 dataset, improvement of RGB+Taylor is +1.1\% and +0.7\% over RGB or Taylor alone, respectively. These experiments demonstrate that the Taylor video is a different but complementary type of input.

\begin{table}[!tbp]
    \centering
    \resizebox{\linewidth}{!}{
    \setlength{\tabcolsep}{4.0pt}
    \begin{tabular}{lcccccc}
        \toprule
        & RGB & \it{Diff.} & \it{Displ.} & \it{Veloc.} & \it{Accel.} & Taylor\\
        \midrule
        Acc(\%) & 59.1 & 53.9 & 61.9 & 62.1 & 59.9 & 65.1\\
        \bottomrule
    \end{tabular}}
    \caption{Evaluations on different combinations of motion concepts on Something-Something v2 with TSM backbone.}
    \label{tab:diff-comb}
\end{table}

\textbf{Performance of Taylor videos and Taylor skeleton sequences under training from scratch}. We first evaluate Taylor video by training networks from scratch on HMDB-51. Results using various 2D/3D models are shown in Table \ref{tab:scratch}. First, compared with fine-tuning with Taylor videos (Table \ref{tab:globaltable}), training from scratch with Taylor videos yields much lower accuracy. Second, under the training-from-scratch setup, Taylor video has similar accuracy compared with RGB videos. 
These results highlight the importance of pre-training for Taylor videos to give superior performance. 

We then present train-from-scratch results on large-scale video datasets and skeleton sequences in Table \ref{tab:large-scale-taylor} and~\ref{tab:taylor-skeleton}, respectively. As shown in Table \ref{tab:large-scale-taylor}, models trained with Taylor videos consistently outperform those trained with RGB videos on all three datasets. We also observe that Taylor-transformed skeleton sequences consistently outperform the original skeleton sequences across various backbones on three large-scale benchmarks (Table \ref{tab:taylor-skeleton}).

\subsection{Further analysis}\label{sec:eval}

\begin{figure}[tbp]
\centering
  \centering\subfigure[Computing a frame]{\label{fig:taylorvsopt}\includegraphics[height=31mm]{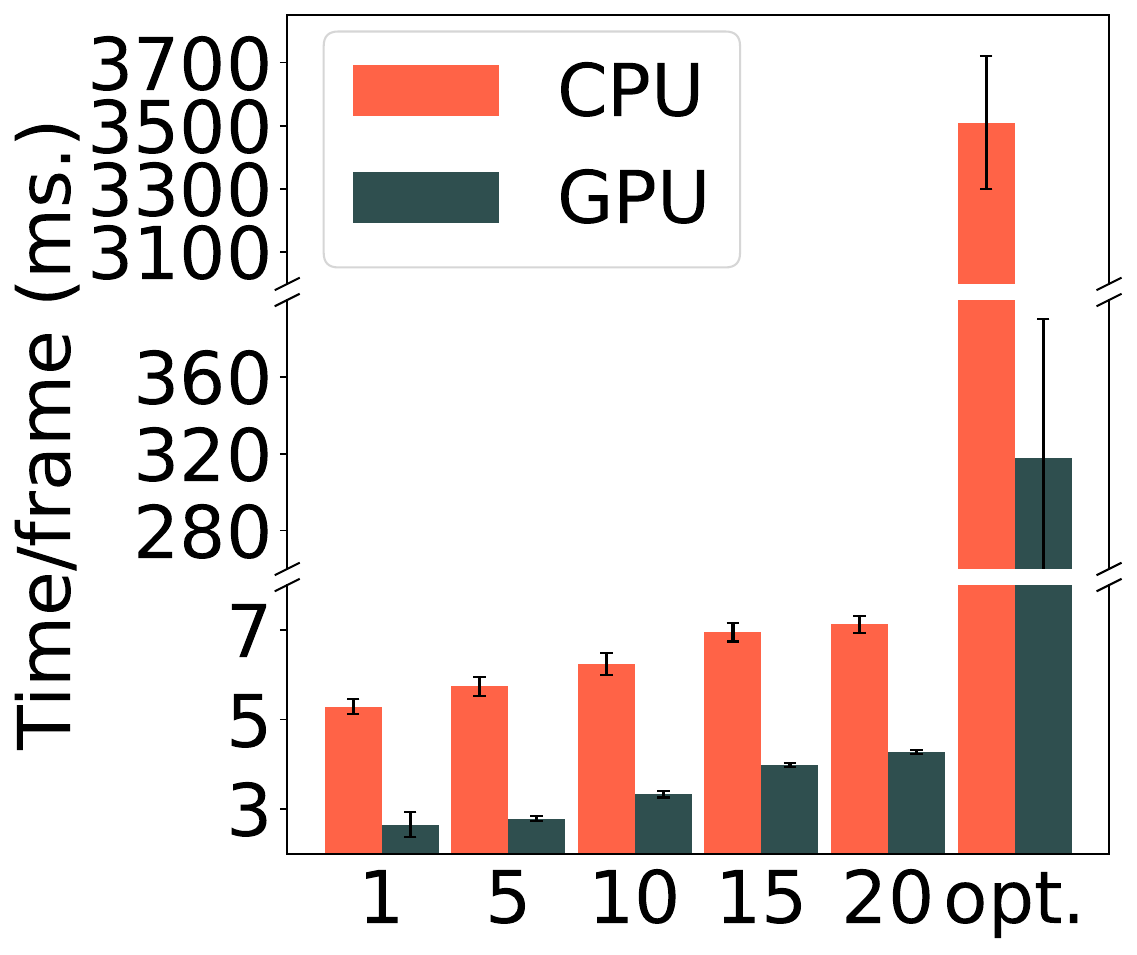}}
\subfigure[Training/Inference]{\label{fig:taylorrgb}\includegraphics[height=31mm]{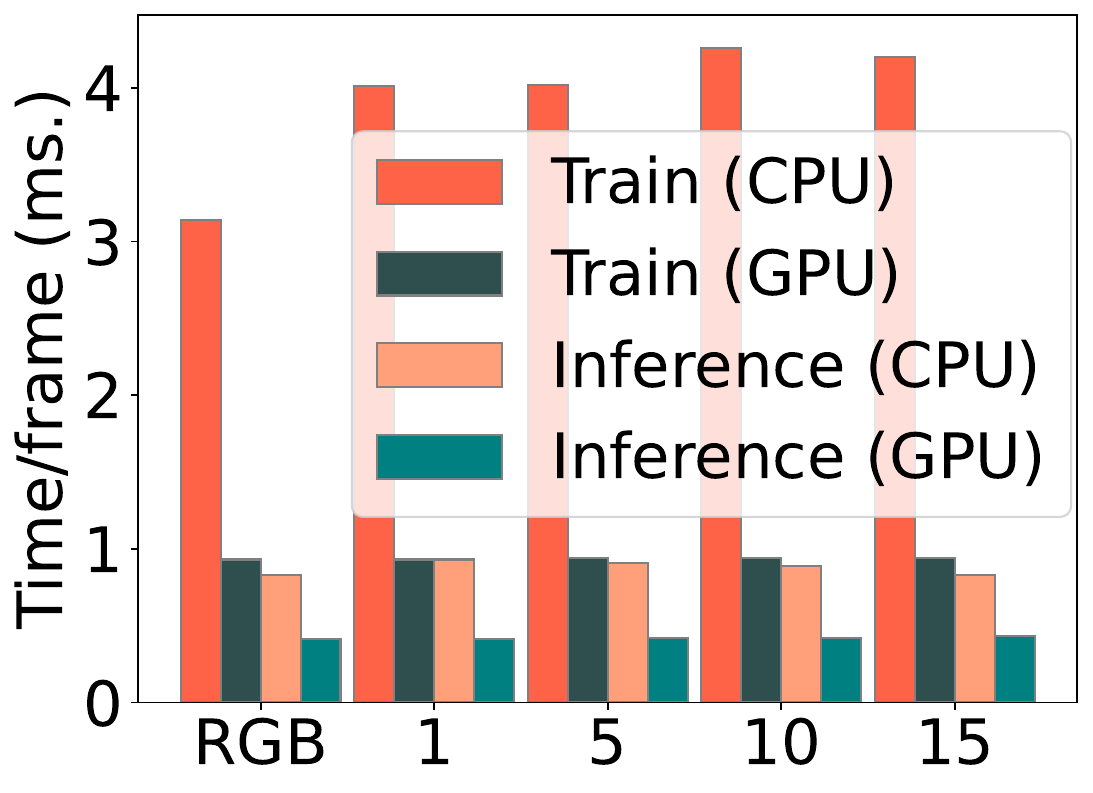}}
\caption{Time cost (milliseconds, ms.) of (a) computing a single Taylor frame (with 1, 5, 10, 15, 20 terms) and optical flow (opt.) and (b) training and testing an RGB frame and Taylor frame (with 1, 5, 10, 15 terms). We use videos from HMDB-51 dataset.}
\end{figure}

\begin{figure*}[!tbp]
    \centering
    \includegraphics[trim=0cm 0.5cm 0cm 0cm, clip=true, width=0.95\linewidth]{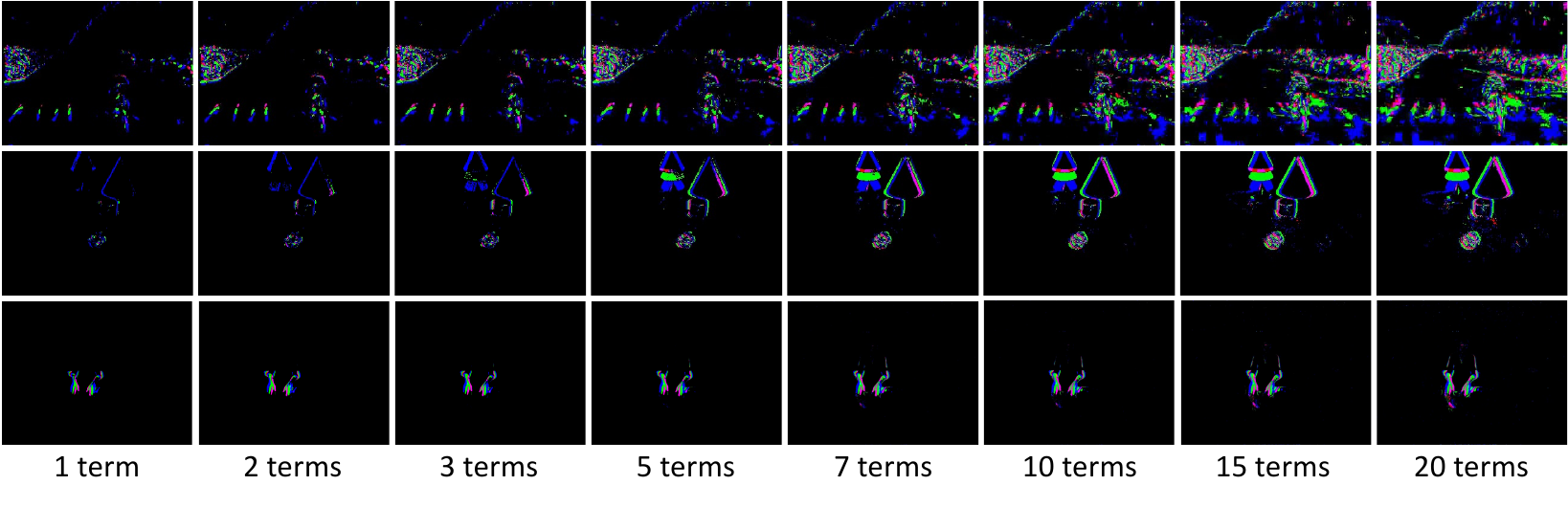}
    \caption{Qualitative impact of the number of terms used in Taylor series on the action \textit{ride bike} in HMDB-51 (top row), a \textit{synthetic action} video of CATER (middle row), and the fine-grained action \textit{place} in MPII Cooking Activity (bottom row). 
    We observe that as the number of terms increases, 
    more motion patterns are captured. In scenarios with a moving camera, such as \textit{ride bike} in the top row, using more terms leads to the inclusion of more background details, such as crosswalk and crowds. For quantitative analysis see Fig. \ref{fig:termsvsacc}.
    }
    \label{fig:compression}
\end{figure*}

\textbf{Computational cost of producing a Taylor frame}. In this and following two experiments, we use 1 NVIDIA Tesla V100 GPU (with 12 CPUs). We use time per frame\footnote{The average time taken to process a single frame: Time per Frame = Total Processing Time / Number of Frames} as the evaluation metric.
In Fig.~\ref{fig:taylorvsopt}, we compare the average time cost of computing a Taylor frame and that of computing an optical flow frame on HMDB-51, where the former cost depends on the length of the temporal block. We demonstrate that the calculation of Taylor frames requires significantly less time than TVL1 optical flow~\cite{tvl1_opt}. 

\textbf{Computational cost of training with Taylor videos}. We assume Taylor videos, RGB videos, and optical flow are all pre-computed. We fine-tuning an I3D model pretrained on Kinetics. From Fig.~\ref{fig:taylorrgb}, we observe the training time for a single Taylor frame is very similar to a single RGB frame, under GPU. On the other hand, Training with taylor videos usually has a cold start: loss drops much more slowly in the beginning (around 20 epochs) but catches up later. A possible reason is that existing architectures are not efficient in reading motion captured by the new Taylor video modality. This cold start affects training speed on small datasets: Taylor videos take around 20 more epochs to converge than RGB videos. But on large datasets such as Kinetics-600, the cold start does not matter due to the overall very long training time: Taylor videos have similar convergence speed with RGB videos, both using about 1200 epochs.

\textbf{Computational cost of inference with Taylor videos}. We also assume pre-computed inputs. We find that Taylor videos take similar time for network processing compared with RGB videos. 
We also note that the efficiency of the Taylor video can be further improved through developing a particular neural network layer dedicated for its computation. This is particularly feasible because the computing of Taylor videos is all composed of in-place matrix operations, which can be easily parallelized.

\textbf{Evaluations on frame differencing and different combinations of motion concepts.} The comparisons using (i) frame differencing maps and (ii) different combinations of motion concepts on Something-Something v2 with TSM are shown in Table~\ref{tab:diff-comb}. We clearly see that frame differencing (\textit{Diff.}) is even worse the RGB video. This is because the difference frames contain lots of noise that negatively affects the network. We observe that incorporating all three motion concepts (Taylor) achieves the best performance compared to using only displacement (\textit{Displ.}), velocity (\textit{Veloc.}), or acceleration (\textit{Accel.}) individually.

\textbf{How many RGB frames are required to compute Taylor video and whether comparing with RGB is fair.} The video trunks used to compute Taylor frames have significant overlaps to ensure motion continuity between Taylor frames. For example, to compute a 16-frame Taylor video, we need 19 and 20 RGB frames if \#terms = 1 and 2, respectively. Note that 2 terms is a good hyperparameter for action recognition. So we are not using much more RGB frames. 

To compute a 16-frame Taylor video, a few more RGB frames, \eg, 19 frames, are needed. This is similar to optical flow: to compute a 16-frame optical flow, 17 RGB frames are needed in the optimal case. In fact, to encode motion, it is necessary to use more RGB frames. We compare models trained with 32-frame Taylor videos and RGB videos of 32 and 64 frames, on HMDB-51 under I3D backbone. The top-1 accuracies are: 65.2\%, 53.4\%, and 49.8\%, respectively, demonstrating the superiority of our method.

\textbf{Impact of the length of temporal blocks}. For the Taylor video defined in this work, the number of frames $T$ in a temporal block is the only hyperparameter, which also determines the degree of Taylor series to be $T\!-\!3$. As discussed before, a degree lower than $T\!-\!3$ will result in under-exploitation of the frames. The evaluation of the impact of $T$ on the HMDB-51 dataset is shown in Fig. \ref{fig:termsvsacc}. 

Interestingly, we observe that the effectiveness of Taylor videos remains stable as the number of frames increases. This suggests a trade-off between capturing long-term motion and introducing noise (see Fig.~\ref{fig:compression}). To be specific, when using fewer frames, \eg, 7 frames and 4 Taylor terms, only the highly dominant motions are encoded, meaning less noisy motions; but to the down side, longer-term motions are not captured. In comparison, if we use 15 or more frames, while long-term motions are computed, using more Taylor terms introduce undesirable noisy motions. 

\begin{table}[!tbp]
    \centering
    \begin{tabular}{lcc}
        \toprule
        & Dataset-level & Action-level \\
        \midrule
        HMDB-51 & 0.87 & 0.68 -- 1.23 \\
        MPII Cooking & 3.04 & 0.93 -- 10.07\\ 
        \bottomrule
    \end{tabular}
    \caption{Compression ratio of Taylor videos on HMDB-51 and MPII on the dataset level and action class level. The number of Taylor terms is 5. A ratio greater than 1 means reduced video sizes, while ratio lower than 1 means increased video sizes. 
    }
    \label{tab:lightweight}
\end{table}

\textbf{Does Taylor videos compress datasets or action classes?} 
We calculate the compression ratio of  RGB videos to Taylor videos (using $\frac{\text{Uncompressed size}}{\text{Compressed size}}$) in different action classes (`action-level') and datasets (`dataset-level') on both HMDB-51 and MPII. 
Results are shown in Fig.~\ref{fig:hmdb51-compression} and Fig.~\ref{fig:mpii-compression} of Appendix and Table \ref{tab:lightweight}. Interestingly, we find compression ratio relatively high for static cameras and `big' motions such as \textit{hug} and \textit{take \& put in oven}, where there exist high background redundancy and dominant motions. In comparison, compression ratio is low for very slow actions, facial movements such as \textit{laugh}, \textit{kiss}, \textit{smile} and \textit{smoke}, and moving cameras. Consequently, converting RGB videos into Taylor videos achieves the highest compression ratio on MPII (static backgrounds and small motion areas). On the contrary, the dataset size even increases for HMDB-51, because of its moving cameras and complex backgrounds. Despite different compression effects on different datasets, our method has consistent accuracy improvement.

\section{Conclusion}
In this paper, we introduce Taylor videos, a new video format for action recognition. Computed through the use of Taylor series for videos, each Taylor video frame captures diverse motion concepts through motion dynamics distillation, effectively eliminating redundancy such as static backgrounds, watermarks, and text captions, while preserving dominant motions for downstream tasks. Taylor videos can be seamlessly integrated into various existing pre-trained RGB-based models, including 2D CNNs, 3D CNNs, and transformer-based architectures, for fine-tuning. 
We show that the Taylor video yields very competitive accuracy compared with the conventional RGB videos, and optical flow, and once combined, produces even better performance. Additionally, we demonstrate that Taylor videos on large-scale datasets and Taylor-transformed skeleton sequences outperform the use of original RGB and skeletons, respectively. 

\pagebreak

\section*{Potential Broader Impact}



From a societal perspective, 
Taylor videos can be adapted to a broad range of areas, such as biology (time-lapse microscopy), medical analysis (endoscopy videos), sports analysis (match footage), physics (motion model), security (surveillance) and many more.

As Taylor videos focus on motions rather than spatial features such as color, our method can provide a higher degree of privacy and security compared to RGB videos. For example, Taylor videos can remove the distinct facial features of individuals within RGB videos (see Sec.~\ref{appendix-face} of supplementary material), allowing data collection and processing to have improved privacy.

\section*{Acknowledgements}
Xiuyuan Yuan conducted this research under the supervision of Lei Wang in the ANU Summer Scholars Program. Lei Wang focused on mathematical analysis and modeling, while Xiuyuan Yuan implemented the code and conducted experiments. Xiuyuan Yuan is supported by a Summer Research Internship provided by the ANU School of Computing. This work is also supported by the NCI Adapter Scheme, with computational resources provided by NCI Australia, an NCRIS-enabled capability supported by the Australian Government. 






\bibliography{example_paper}
\bibliographystyle{icml2024}

\newpage
\appendix
\onecolumn

\section{Proof of Equivalence}
\label{sec:appendix-proof}

Below we provide the proof that Eq.\eqref{eq:tensor} is equivalent to Eq.\eqref{eq:sum-motion-concept}:
\begin{align}
    & \!\!\!\!\!t(\tF)\!=\!\sum_{k=0}^\infty \frac{t^{(k)}(\widetilde{\tF})}{k!}\!\odot\frac{1}{T}\!\sum_{\tau=1}^{T}(\tF-\widetilde{\tF})^{\circ k}_{:,:, \tau}\!=\frac{1}{T}\!\sum_{\tau=1}^{T} \sum_{k=0}^\infty \frac{t^{(k)}(\widetilde{\tF})}{k!}\!\odot\!(\tF-\widetilde{\tF})^{\circ k}_{:,:, \tau} \nonumber \\
    & \quad \!=\frac{1}{T}\!\sum_{\tau=1}^{T} \sum_{k=0}^\infty\!\frac{t^{(k)}(\widetilde{\tF})}{k!}\!\odot\!([\mF_1\!-\!\mF_1, \mF_2\!-\!\mF_1, \cdots, \mF_\tau\!-\!\mF_1,\cdots,\mF_{T}\!-\!\mF_1])^{\circ k}_{:,:, \tau} \nonumber\\
    & \quad \!=\frac{1}{T}\!\sum_{\tau=1}^{T} \sum_{k=0}^\infty\!\frac{t^{(k)}(\widetilde{\tF})}{k!}\!\odot\![(\mF_1\!\!-\!\!\mF_1)^{\circ k}, (\mF_2\!\!-\!\!\mF_1)^{\circ k}, \cdots, \nonumber \\
    & \quad \quad \quad \quad \quad \quad \quad \quad \quad \quad \quad(\mF_{\tau}\!\!-\!\!\mF_1)^{\circ k}, \cdots, (\mF_{T}\!\!-\!\!\mF_1)^{\circ k}]_{:,:, \tau} \nonumber\\
    & \quad \!=\frac{1}{T}\!\sum_{\tau=1}^{T} \sum_{k=0}^\infty\!\left[\frac{t^{(k)}(\widetilde{\tF})}{k!}\!\odot\!(\mF_1\!\!-\!\!\mF_1)^{\circ k}, \frac{t^{(k)}(\widetilde{\tF})}{k!}\!\odot\!(\mF_2\!\!-\!\!\mF_1)^{\circ k}, \cdots, \right. \nonumber \\
    & \quad \quad \quad \quad \quad \quad \quad \left. \frac{t^{(k)}(\widetilde{\tF})}{k!}\!\odot\!(\mF_{\tau}\!\!-\!\!\mF_1)^{\circ k}, \cdots, \frac{t^{(k)}(\widetilde{\tF})}{k!}\!\odot\!(\mF_{T}\!\!-\!\!\mF_1)^{\circ k}\right]_{:,:, \tau}  \nonumber\\
    & \quad \!=\frac{1}{T}\! \sum_{\tau=1}^{T} \sum_{k=0}^\infty\!\frac{t^{(k)}(\widetilde{\tF})}{k!}\!\odot\!(\mF_{\tau}\! - \!\mF_1)^{\circ k}. 
    \label{eq:proof}
\end{align}

Note that Eq.\eqref{eq:proof} is equivalent to Eq.\eqref{eq:sum-motion-concept} under the following conditions: (i) $\mF_1$ is the first frame of temporal block $\tF\!=\![\mF_1, \mF_2, \cdots, \mF_{T}]\!\in\!\mbr{H\!\times\!W\!\times\!T}$, and $\widetilde{\tF}\!=\![\mF_1, \mF_1, \cdots, \mF_1]\!\in\!\mbr{H\!\times\!W\!\times\!T}$ denotes the generated dummy temporal block with no motion, (ii) both $f^{(k)}(\mF_1)$ and $t^{(k)}(\widetilde{\tF})$ captures the motions and dynamics within the temporal block. It is important to note that while the inputs to $f$ and $t$ differ, both functions capture the same motions. Specifically, $f$ captures motions until frame $\mF_{\tau}$, whereas $t$ captures motions within the given temporal block.

\section{Algorithm for Efficient Implementation}
\label{sec:alg}
Algorithm~\ref{alg:taylor} shows the efficient implementation of the Taylor video.

\begin{algorithm}[!tbp]
   \caption{Efficient implementation of Taylor video}
   \label{alg:taylor}
\begin{algorithmic}
   \STATE {\bfseries Input:} Grayscale video $\tF\in\!\mbr{H\!\times\!W\!\times\!\mathcal{T}}$, total number of terms $K$, temporal sliding window size $T$ (step size is 1)
   \IF{$T\!-\!3 < K$}
   \STATE Print `The given temporal block length $T$ is not enough to compute $K$ terms.'
    \ELSE
    \STATE {Create $\tM\!\in\!\mbr{H\!\times\!W\!\times\!3\!\times\!N}$ ($N\!=\!\mathcal{T}\!-\!T\!+\!1$) for storing Taylor video}
   \FOR{$i=1$ {\bfseries to} $\mathcal{T}\!-\!T\!+\!1$}
   \STATE {Get the $i$th video temporal block $\tF^i\!\in\!\mbr{H\!\times\!W\!\times\!T}$}
   \STATE {Create a temporary copy: $\mathbf{\Delta}_\text{temp}^i = \tF^i$}
   \STATE {Create a static video temporal block $\widetilde{\tF}^i\!\in\!\mbr{H\!\times\!W\!\times\!T}$} by duplicating the first frame of $\tF^i$
   \STATE Create $\tD^i$ to store different orders of frame differencing maps
   \FOR{$j=1$ {\bfseries to} $K\!+\!2$}
   \STATE {Compute frame differencing maps: $\mathbf{\Delta}_j^i\!=\!\mathbf{\Delta}_\text{temp}^i[:, :, 1\!:] - \mathbf{\Delta}_\text{temp}^i[:, :, :\!-1]$}
   \STATE {Save the 1st frame differencing map: $\tD^i[:, :, j\!-\!1]\!=\!\mathbf{\Delta}_j^i[:, :, 0]$}
   \STATE {Save temporary frame differencing maps: $\mathbf{\Delta}_\text{temp}^i = \mathbf{\Delta}_j^i \in \mbr{H\!\times\!W\!\times\!(T-j)}$}
   \ENDFOR
    \STATE {Initialise $\tM^i_{d}, \tM^i_{v}, \tM^i_{a}\!=\!\mathbf{0}\!\in\mbr{H\!\times\!W\!\times\!T}$}
    \STATE {Compute $\tX^i = \tF^i\!-\!\widetilde{\tF}^i$}
   \FOR{$k=0$ {\bfseries to} $K$}
   \STATE {$\tM^i_{d} \!=\!\tM^i_{d}\!+\!\frac{\tD^i[:, :, k]}{k!} \odot (\tX^i)^{\circ k}$}
   \STATE {$\tM^i_{v} \!=\!\tM^i_{v}\!+\!\frac{\tD^i[:, :, k\!+\!1]}{k!} \odot (\tX^i)^{\circ k}$}
   \STATE {$\tM^i_{a} \!=\!\tM^i_{a}\!+\!\frac{\tD^i[:, :, k\!+\!2]}{k!} \odot (\tX^i)^{\circ k}$}
   \ENDFOR
   \STATE Get the $i$th Taylor frame: $\mM^i\!=\![\tM^i_{d}.\text{mean}(2); \tM^i_{v}.\text{mean}(2); \tM^i_{a}.\text{mean}(2)] \in \mbr{H\!\times\!W\!\times\!3}$
   \STATE Form a Taylor video: $\tM[:, :, :, i]\!=\!\mM^i$
   \ENDFOR
   \ENDIF
   \STATE {\bfseries Output:} Taylor video $\tM\!\in\!\mbr{H\!\times\!W\!\times\!3\!\times\!N}$
\end{algorithmic}
\end{algorithm}

\section{Action-level Compression}
\label{sec: appendix-action-compression}

\begin{figure}[tbp]
    \centering
    \includegraphics[width=0.6\linewidth]{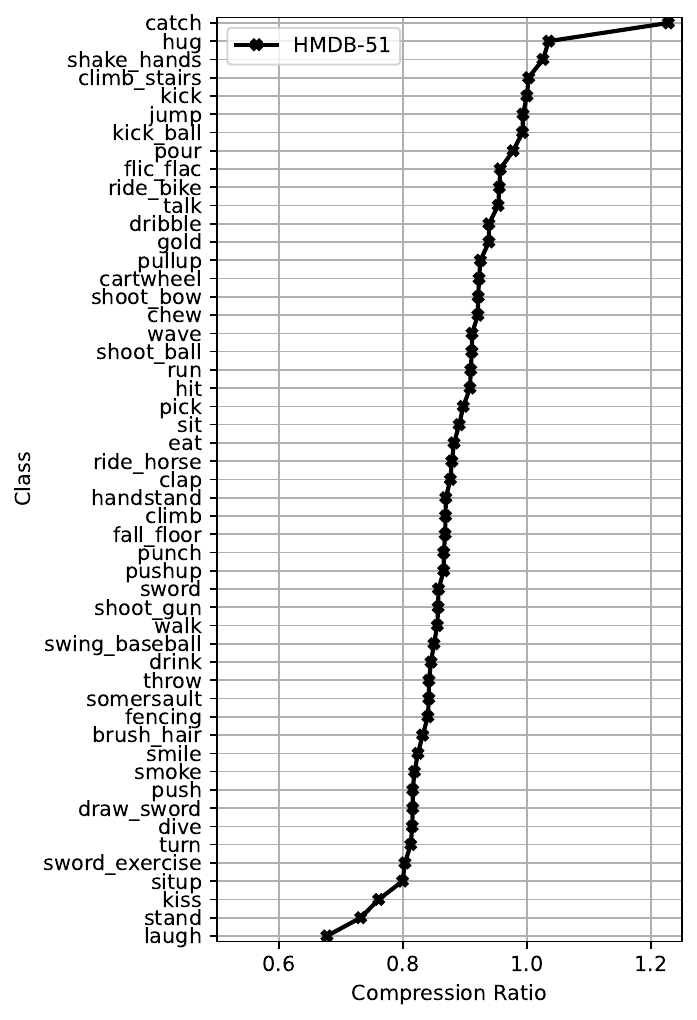}
    \vspace{-0.5cm}
    \caption{Compression ratio per action on HMDB-51.}
    \label{fig:hmdb51-compression}
\end{figure}

\begin{figure}[tbp]
    \centering
    \includegraphics[width=0.7\linewidth]{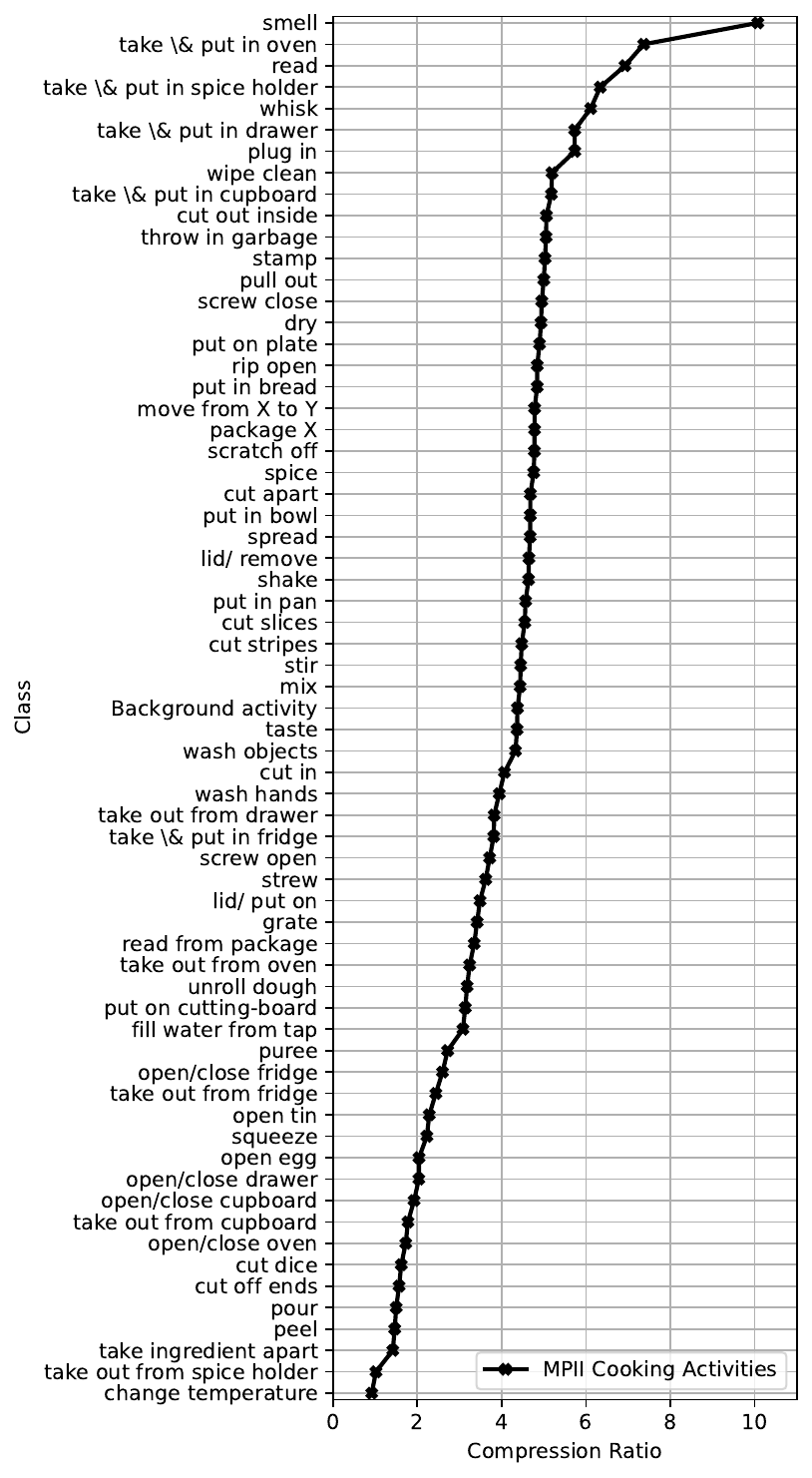}
    \vspace{-0.5cm}
    \caption{Compression ratio per action on MPII Cooking Activities.}
    \label{fig:mpii-compression}
\end{figure}

Below we provide visualisations on action-level compression ratio on both HMDB-51 and MPII Cooking Activity.

Fig.~\ref{fig:hmdb51-compression} shows the results for HMDB-51. It is noteworthy that, on average, the action \textit{catch} exhibits the highest compression ratio compared to other actions. Interestingly, actions such as \textit{smile}, \textit{kiss}, and \textit{laugh} demonstrate the lowest compression ratios. This can be attributed to the fact that these actions primarily occur around facial regions, making it challenging to capture their dominant motions compared to actions like \textit{catch}, \textit{hug} and \textit{kick}. 

Another potential factor contributing to this observation is the lower resolution and inherent noise present in the HMDB-51 dataset. Extracting dominant motions from videos with these characteristics proves to be a challenging task.

Fig.~\ref{fig:mpii-compression} shows the results for MPII Cooking Activity. As shown in the figure, most actions have more than $2\times$ compression ratio on average, it shows that fine-grained action recognition dataset indeed has much  redundancy. We also notice that some actions such as \textit{change temperature} and \textit{take out from spice holder} have lower compression ratio. The possible reason behind this phenomenon is that these actions are too tiny and sometimes even smaller than the background noisy patterns, hence these noises become the dominant motions.

\section{Taylor Frames for Face Videos}

\label{appendix-face} 

We use videos from the following two datasets to compute the Taylor videos.

\noindent\textbf{Celeb-DF (v2)}~\cite{Celeb_DF_cvpr20} dataset contains real and DeepFake synthesized videos having similar visual quality on par with those circulated online. The Celeb-DF (v2) dataset is greatly extended from the previous Celeb-DF (v1), which only contains 795 DeepFake videos. To date, Celeb-DF includes 590 original videos collected from YouTube with subjects of different ages, ethic groups and genders, and 5639 corresponding DeepFake videos.

\noindent\textbf{YouTube Faces}~\cite{5995566} is a database of face videos designed for studying the problem of unconstrained face recognition in videos.
The data set contains 3,425 videos of 1,595 different people. All the videos were downloaded from YouTube. An average of 2.15 videos are available for each subject. The shortest clip duration is 48 frames, the longest clip is 6,070 frames, and the average length of a video clip is 181.3 frames. 

\begin{figure}[tbp]
    \centering
    \includegraphics[width=\linewidth]{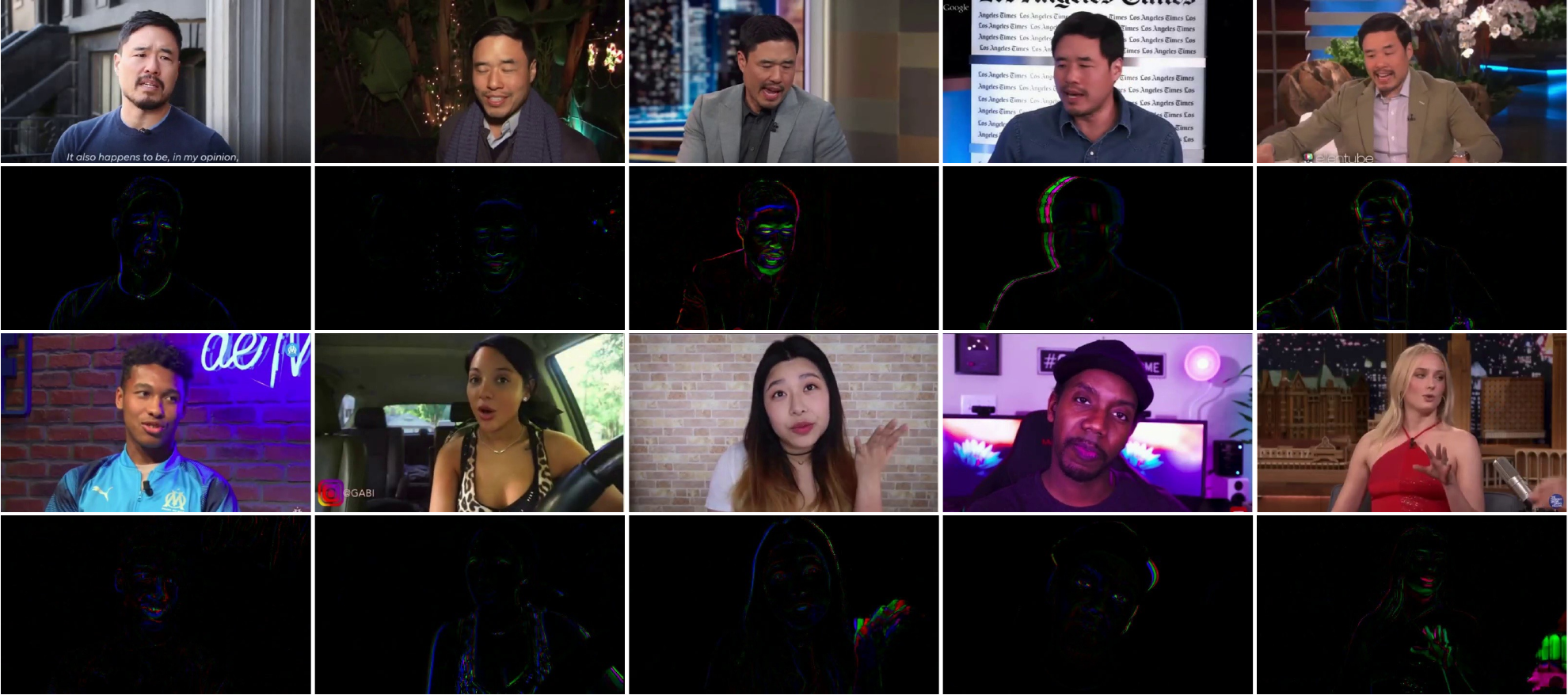}
    \vspace{-0.5cm}
    \caption{We use videos from ({\it Top two rows}) Celeb-DF (v2) and ({\it Bottom two rows}) YouTube Faces to show that Taylor videos are able to remove distinct facial features of individuals compared to RGB videos. This allows the data collection and processing to have improved privacy.}
    \label{fig:face}
\end{figure}

Fig.~\ref{fig:face} shows that Taylor videos can remove the distinct facial features of individuals within RGB videos. This allows the data collection and processing to have improved privacy.

\end{document}